\documentclass[12pt]{article}

\usepackage{microtype}
\usepackage{graphicx}
\usepackage{subfigure}
\usepackage{booktabs} 

\usepackage{hyperref}       
\usepackage{url}            
\usepackage{booktabs}       
\usepackage{amsfonts}       
\usepackage{nicefrac}       
\usepackage{microtype}      
\usepackage{amsthm}

\usepackage{amsmath}

\usepackage{bbm}
\usepackage[margin=1in]{geometry}

\usepackage{thmtools}
\usepackage{lipsum}

\declaretheoremstyle[
spaceabove=\topsep, spacebelow=\topsep,
headfont=\normalfont\bfseries,
notefont=\bfseries, notebraces={}{},
bodyfont=\normalfont\itshape,
postheadspace=0.5em,
name={\ignorespaces},
numbered=no,
headpunct=.]
{mystyle}
\declaretheorem[style=mystyle]{named}

\DeclareMathOperator{\Ex}{\mathbb{E}}
\DeclareMathOperator{\Prob}{\mathbb{P}}

\DeclareMathOperator{\Hp}{\mathcal{H}}

\DeclareMathOperator*{\argmax}{arg\,max}

\newtheorem{theorem}{Theorem}
\newtheorem{lemma}[theorem]{Lemma}

\newtheorem{corollary}[theorem]{Corollary}

\usepackage{setspace}
\usepackage{hyperref}

\usepackage[numbers]{natbib}
\title{Asymptotic Convergence of Thompson Sampling}
\author{ Cem Kalkanl\i \\ 
	cemk@stanford.edu \\
	 Stanford University\\
	Stanford, CA 94305, USA
	\and
	 Ayfer \"{O}zg\"{u}r \\ 
	 aozgur@stanford.edu \\
	 Stanford University\\
	Stanford, CA 94305, USA}
\date{}
\begin{document}
\maketitle

\begin{abstract}
Thompson sampling has been shown to be an effective policy across a variety of online learning tasks. Many works have analyzed the finite time performance of Thompson sampling, and proved that it achieves a sub-linear regret under a broad range of probabilistic settings. However its asymptotic behavior remains mostly underexplored. In this paper, we prove an asymptotic convergence result for Thompson sampling under the assumption of a
sub-linear Bayesian regret, and show that the actions of a Thompson
sampling agent  provide a strongly
consistent estimator of the optimal action. Our results rely on the martingale structure inherent in Thompson sampling.    
\end{abstract}

\section{Introduction}\label{intro}

In the multi-armed bandit problem, the agent repeatedly selects an action, a.k.a an arm, from an action set, and as a result observes a random outcome, e.g. a reward. The agent aims to maximize the total expected reward. Since the observations are noisy and the underlying system is unknown to the agent, she has to balance between the two factors: the amount of information gathering for a possible high future reward (exploration), and the maximization of the immediate reward according to the current knowledge about the system (exploitation). In one of the earliest versions of this multi-armed bandit problem, \citet{ts} proposed an algorithm known as Thompson sampling where at each turn the agent randomly selects an action according to its likelihood of being optimal. Recently \citet{cl11} showed that Thompson sampling is an effective heuristic, in some cases performing better than the known Upper Confidence Bound (UCB) algorithms, when it comes to the areas such as the display advertising, and the news article recommendation. Its state-of-the-art empirical performance \cite{cl11,dw18}, and its ease of implementation, which does not require any tuning unlike a UCB algorithm, sparked an interest in Thompson sampling, and many works provided theoretical performance guarantees for the Thompson sampling policy. Russo and Van Roy \cite{rv14,rv16} established a bound of $\tilde{O}(\sqrt{T})$\footnote{$\tilde{O}$ hides poly-logarithmic factors.}, where $T$ stands for the duration of the experiment, on the Bayesian regret of the Thompson sampling policy for almost all known multi-armed bandit problems, while \citet{km12,ag17} analyzed the expected number of times the Thompson sampling agent played a sub-optimal arm, and found it to be no more than $O(\log(T))$ for the multi-armed bandit problems with independent arms, and Beta priors.

In this paper, we focus on a different aspect of Thompson sampling. We aim to understand if Thompson sampling is an effective strategy for asymptotic exploration. We ask the following question: can Thompson sampling be used to obtain a strongly consistent estimator for the optimal action? In other words, if run for sufficiently large number of steps can the Thompson sampling algorithm be used to discover the optimal action? Even when there is a strict separation between the reward achieved by the optimal action and that
achieved by any of the other actions, it is not guaranteed that the actions selected by a general algorithm, which achieves a sub-linear regret, will converge to the optimal one even in probability. Additionally, in many settings, e.g. the canonical case of the Gaussian linear bandit, such strict separation does not exist. In this work, we show that a sub-linear Bayesian regret for Thompson sampling leads to a strongly consistent estimator for the optimal action  if the action set is countable. Our proofs rely on the martingale property regarding the action selection process of Thompson sampling. To the best of our knowledge, the strongly consistent estimators has not been considered for Thompson sampling in the previous literature.

\section{Problem Setup}

\subsection{The Multi-Armed Bandit Problem}\label{mab}

We consider a multi-armed bandit problem in which an agent chooses an action $A_t$ from a countable action set $\mathcal{A}=\{a_1,a_2,...\}$, which admits an ordering, at each time step $t\in\mathbb{Z}^+$, and observes a reward $Y_{A_t,t}$:
\begin{equation*}
	Y_{A_t,t}=g(\theta^*,A_t,W_t)
\end{equation*}
where $\theta^*$ is the random system variable that takes the values in the parameter set $\Theta$, $W_t\in\mathbb{R}$ for $t\in\mathbb{Z}^+$ is the noise variable, and $g$ is a function from $\Theta\times\mathcal{A}\times\mathbb{R}$ to $\mathbb{R}$. The random variables $\theta^*$, and $W_t$'s are mutually independent, and unknown to the agent. 

We assume that
\begin{align}
	&\Ex[\sup_{a\in\mathcal{A}}|Y_{a,t}|]<\infty,\text{ and}\label{e1}\\
	&\Ex[Y_{a,t}|\theta^*]=f(\theta^*,a)\text{ almost surely (a.s.)}\label{e2}
\end{align}  
for any $t\in\mathbb{Z}^+$, and $a\in\mathcal{A}$ where $f$ is a function from $\Theta\times\mathcal{A}$ to $\mathbb{R}$. Note that \eqref{e1}, and \eqref{e2} implies
\begin{equation}
	\Ex[\sup_{a\in\mathcal{A}}|f(\theta^*,a)|]<\infty\label{e3}.
\end{equation}
Since $f(\theta,a)$ is the average reward the agent receives when she chooses the action $a\in\mathcal{A}$ under the true parameter $\theta\in\Theta$, we define the optimal action $A^*:\Theta\rightarrow\mathcal{A}$ as the function such that
\begin{equation}
	A^*(\theta)=\argmax_{a\in\mathcal{A}}f(\theta,a).\label{e4}
\end{equation}
Note that there might be more than one candidate for the optimal action $A^*$. To resolve this dispute over uniqueness of the optimal action, we use the natural ordering of the actions in $\mathcal{A}$ such that 
\begin{equation}
	A^*(\theta)=a_{i_{min}}\label{e5}
\end{equation}
where 
\begin{equation}
	i_{min}=\min\{i|a_i\in\mathcal{A},\text{ and }f(\theta,a_i)\geq f(\theta,a)\quad \forall a\in\mathcal{A}\}\label{e6}
\end{equation}
for any $\theta\in\Theta$. Under the preceding construction, we denote $\Theta_i$ as
\begin{equation}
	\Theta_i=\{\theta\in\Theta|A^*(\theta)=a_i\}\label{e7}
\end{equation}
for any $a_i\in\mathcal{A}$. It is clear that $A^*$ defined in \eqref{e5}, and \eqref{e6} satisfies \eqref{e4}. 

We define the corresponding reward $Y_{A^*,t}$ when the optimal action $A^*(\theta^*)$ is played for any $t\in\mathbb{Z}^+$ as
\begin{equation*}
	Y_{A^*,t}=g(\theta^*,A^*(\theta^*),W_t).
\end{equation*}

Since the agent can only act causally at any given time $t\in\mathbb{Z}^+$, she can only use the information present in $\Hp_{t-1}$ where
\begin{equation*}
\Hp_{t-1}=\{A_1,Y_{A_1,1},...,A_{t-1},Y_{A_{t-1},t-1}\}	
\end{equation*}
such that $\Hp_0=\emptyset$, and $\Hp_{\infty}=\cup_{t=1}^{\infty} \Hp_t$. Thus, the agent follows a policy $\pi=(\pi_1,\pi_2,...)$, which describes the rules to choose an action at each time step, such that for any $t\in\mathbb{Z}^+$ $\pi_t$ is a map from $\Hp_{t-1}$ to a probability distribution on $\mathcal{A}$:
\begin{equation*}
	\Prob(A_t\in B|\Hp_{t-1})=\pi_t(\Hp_{t-1})(B)\quad\text{a.s.}
\end{equation*}
for any $B\subseteq \mathcal{A}$.

	Finally, we define $N_{B,T}$ as the number of times the agent visits the set $B\subseteq \mathcal{A}$ in the first $T$ time steps such that
\begin{equation*}
N_{B,T}=\sum_{t=1}^{T}I_B(A_t),
\end{equation*}
where $I_{B}(\cdot)$ denotes the indicator function whose domain is $B$.
 
\subsection{Thompson Sampling}\label{TS-MG}

In this paper, we consider the Thompson sampling policy where the agent who follows it selects the action $A_t$ according its likelihood of being optimal given the earlier observations $\Hp_{t-1}$. This relation leads to the following for any $B\subseteq\mathcal{A}$:
\begin{equation}
	\Prob(A^*(\theta^*)\in B|\Hp_{t-1})=\Prob(A_t\in B|\Hp_{t-1})\quad\text{a.s.}\label{e8}
\end{equation} 
Here, \eqref{e8} summarizes the operation of Thompson sampling. The left-hand side of \eqref{e8} captures the posterior distribution of $A^*(\theta^*)$, which is a random variable since it is the optimal action corresponding to the random system variable $\theta^*$, given past observations $\Hp_{t-1}$. This equation simply states that at each time step $t$, the Thompson sampling agent draws an action $A_t$ randomly from this posterior distribution, i.e. the probability with which the Thompson sampling agent chooses an action is equal to the probability that this action is optimal given the past observations. This operation is naturally equivalent to first drawing from the posterior distribution of $\theta^*$ given $\Hp_{t-1}$ and then playing the optimal action corresponding to this sample. 

As a result, the relationship given in \eqref{e8} leads to the $\Hp_t$-adapted random variables $\{\Prob(A_{t+1}\in B|\Hp_{t})\}_{t=0}^{\infty}$ becoming a martingale sequence, since for any $s\geq t\geq 0$:
\begin{align}
	\Ex[\Prob(A_{s+1}\in B|\Hp_{s})|\Hp_t]&=\Ex[\Prob(A^*(\theta^*)\in B|\Hp_{s})|\Hp_t]\label{m1}\\
	&=\Prob(A^*(\theta^*)\in B|\Hp_{t})\label{m2}\\
	&=\Prob(A_{t+1}\in B|\Hp_{t})\quad\text{a.s.}\label{m3}
\end{align}
where \eqref{m1} follows from \eqref{e8}. Also $\Hp_t$ being a subset of $\Hp_s$ leads to \eqref{m2}. Similarly, \eqref{m3} follows from \eqref{e8}. Consequently, this martingale structure provides the following convergence result: 
\begin{align}
	\lim_{t\rightarrow\infty}\Prob(A_t\in B|\Hp_{t-1})&=\lim_{t\rightarrow\infty}\Prob(A^*(\theta^*)\in B|\Hp_{t-1})\label{e9}\\
	&=\Prob(A^*(\theta^*)\in B|\Hp_{\infty})\quad\text{a.s.}\label{e10}
\end{align}
where \eqref{e9} directly follows from \eqref{e8}. To prove \eqref{e10}, we use the next theorem which is related to the martingales:

\begin{named}[\textbf{Theorem 4.6.8} of \cite{d19}]\label{thm1}
	Suppose $\Ex[|X|]<\infty$ for some random variable $X$. As $t\rightarrow\infty$,
	\begin{equation*}
			\Ex[X|\Hp_t]\rightarrow\Ex[X|\Hp_{\infty}]\quad\text{a.s. and in }L^1.
	\end{equation*}
\end{named}
	If we let $I_{B}(A^*(\theta^*))$ be $X$ in this theorem, and note that $\Ex[I_{B}(A^*(\theta^*))|\Hp_{t-1}]=\Prob(A^*(\theta^*)\in B|\Hp_{t-1})$ a.s., we achieve \eqref{e10}.

	\subsection{Consistent Estimator for $A^*(\theta^*)$}
	Let $B$ be any subset of $\mathcal{A}$. We define any $\Hp_t$-adapted sequence $\{E_t(B)\}_{t=1}^{\infty}$ as a consistent estimator for $I_{B}(A^*(\theta^*))$ if
	\begin{equation*}
		\lim_{t\rightarrow\infty}\Prob(|E_t(B)-I_{B}(A^*(\theta^*))|>\epsilon)=0
	\end{equation*} 
	for any $\epsilon>0$. In the case that this convergence happens almost surely such that
	\begin{equation*}
		\lim_{t\rightarrow\infty}E_t(B)=I_{B}(A^*(\theta^*))\quad\text{a.s.}
	\end{equation*}
	then we call the estimator $\{E_t(B)\}_{t=1}^{\infty}$ strongly consistent. 
	
	In the following section, we will construct an estimator for the optimal action $A^*(\theta^*)$ which is strongly consistent if Thompson sampling achieves a sub-linear Bayesian regret.
	\section{Main Results}\label{conv}
	In this section, we state our main results. Note that all results assume that the agent follows the Thompson sampling policy.
	
	\begin{theorem}\label{thm2}
		Let $B\subseteq\mathcal{A}$. Then we have 
		\begin{equation*}
		\lim_{T\rightarrow\infty}\frac{1}{T}N_{B,T}=\Prob(A^*(\theta^*)\in B|\Hp_\infty)\quad\text{a.s.}
		\end{equation*}
	\end{theorem}
	We provide the proof of this theorem in Section \ref{pr1}.

	Note that Theorem \ref{thm2} applies to any bandit setting described in Section \ref{mab} and only relies on the martingale property of Thompson sampling, \eqref{m3}, in its proof. Therefore it holds in general for Thompson sampling applied to any multi-armed bandit setting. In the next theorem, we show that when Thompson sampling achieves a sublinear Bayesian regret, the term $\Prob(A^*(\theta^*)\in B|\Hp_\infty)$ in Theorem \ref{thm2} can be replaced by the indicator random variable $I_B(A^*(\theta^*))$.

	\begin{theorem}\label{thm4}
		Suppose the Thompson sampling policy achieves a sub-linear Bayesian regret, meaning that
		\begin{equation}
			\lim_{T\rightarrow\infty}\frac{1}{T}\sum_{t=1}^T \Ex[Y_{A^*,t}-Y_{A_t,t}]=0,\label{e16}
		\end{equation}
		then for any $B\subseteq\mathcal{A}$ we have
		\begin{equation}
			\Prob(A^*(\theta^*)\in B|\Hp_\infty)=I_B(A^*(\theta^*))\quad\text{a.s.},\label{e17}
		\end{equation}
		and
		\begin{equation}
			\lim_{t\rightarrow\infty}\Prob(A_t\in B|\Hp_{t-1})=I_B(A^*(\theta^*))\quad\text{a.s.}.\label{e18}
		\end{equation}

	\end{theorem}
The proof of this theorem is given in Section \ref{pr2}.

	As we have mentioned earlier in Section \ref{intro}, the condition stated in \eqref{e16} is satisfied for a wide range of multi-armed bandit problems including canonical models such as the linear bandit and the logistic bandit. See for example \cite{rv14,rv16},  which prove Bayesian regret bounds of order $\tilde{O}(\sqrt{T})$ on the expected cumulative regret of Thompson sampling for a wide range of multi-armed bandit problems.
	
	 Equation \eqref{e18} of Theorem \ref{thm4} shows that if Thompson sampling achieves a sub-linear Bayesian regret, then the probability of sampling an action converges to 0 or 1 depending on this action being optimal. In other words, if the Bayesian regret achieved by Thompson sampling is sub-linear, the theorem ensures that as $t$ gets large the probability of sampling the optimal action will converge to $1$ and the probability of sampling a suboptimal action will converge to $0$. This result may seem intuitive at first as one may be inclined to think that a sublinear regret should implicitly imply convergence to the optimal action. However, it is not true in general, and the result in Theorem~\ref{thm4} critically relies on the martingale structure of Thompson sampling. We next provide an example to illustrate this point. Consider a policy $\pi^1$ that achieves a regret of $O(\sqrt{T})$. Using $\pi^1$, we construct a new policy $\pi^2$ as follows. Let $\pi^2$ play a fixed predetermined action $a\in\mathcal{A}$ on time steps $t=i^2$ for $i\in\mathbb{Z}^+$ and play actions according to $\pi^1$ in the remaining times steps, i.e. $\pi^2$ is a combination of a constant policy and $\pi^1$ where both constituent policies ignore the observations of the other. With this construction, it is easy to see that the regret of $\pi^2$ is $O(\sqrt{T})$. This is because over a horizon of $T$, the constant policy is played in less than $\sqrt{T}$ steps; therefore, its contribution to the regret is bounded by $O(\sqrt{T})$. As for $\pi^1$, it is played less than $T$ times, and therefore given our initial assumption its contribution to the regret is bounded by $O(\sqrt{T})$. However, even though this analysis shows that $\pi^2$ achieves a sub-linear regret, $O(\sqrt{T})$ to be precise, $\pi^2$ plays a fixed sub-optimal action infinitely often. As a result, we cannot ensure that the actions of $\pi^2$ will be in a small neighborhood of the optimal action even as $T$ gets large. Here, the probability that the action taken by $\pi^2$ is in a certain neighborhood of the optimal action does not converge, which shows that such convergence is not guaranteed for any policy that achieves sub-linear regret.

	\begin{corollary}\label{cor}
		Assume that the Thompson sampling policy satisfies \eqref{e16},
		then for any $B\subseteq\mathcal{A}$ we have
		\begin{equation}
			\lim_{T\rightarrow\infty}\frac{1}{T}N_{B,T}=I_B(A^*(\theta^*))\quad\text{a.s.}\label{co1}
		\end{equation}
		\begin{proof}
			Combining Theorem \ref{thm2} with \eqref{e17} of Theorem \ref{thm4} gives the desired result. 
		\end{proof}
	\end{corollary}

	This corollary provides a strongly consistent estimator $\{\frac{1}{T}N_{B,T}\}_{T=1}^\infty$, which is adapted to $\Hp_T$, for the optimal action. Note that we can construct the estimator by simply observing the actions taken by the Thompson sampling agent, and do not require any knowledge about the inner workings of the multi-armed bandit setup. Thus, this corollary allows an external observer to estimate the optimal action by simply observing the actions of the agent over a large horizon.
	
	On the other hand, \citet{km12,ag17} proved that 
	\begin{equation}
	\limsup_{T\rightarrow\infty}\frac{\Ex[N_{\{a\},T}|\theta^*]}{\log(T)}<\infty\quad\text{a.s.}\label{h1}
	\end{equation}
	if $f(\theta^*,a)<f(\theta^*,A^*(\theta^*))$ in the case that the setup is the Beta-Bernoulli bandit with independent arms. Although \eqref{h1} provides a sharper convergence rate when $A^*(\theta^*)\not\in B$ compared with \eqref{co1}, it is still defined in terms of a conditional expectation unlike the result in \eqref{co1}, which is true for almost surely all sample paths. Also \eqref{h1} differs from Corollary \ref{cor} in terms of applicability since the former applies only to a specific bandit problem while the latter remains true for wide range of multi-armed bandit problems with known sub-linear Bayesian regrets bounds \cite{rv14,rv16}.

\section{Conclusion}

By using the martingale structure present in Thompson sampling and the Bayesian regret bounds available in \cite{rv14,rv16}, we proved that the Thompson sampling agent can accurately predict the optimal action if the experiment is run for sufficiently long time. Building on top of this convergence result, we also constructed a strongly consistent estimator for the optimal action which only depends on the actions taken by the agent. As far as we know, this type estimator is the first of its kind in the literature with regards to Thompson sampling.

\section{Proofs}
	In this section, we provide the proofs of Theorem \ref{thm2}, and \ref{thm4}.
	\subsection{Proof of Theorem \ref{thm2}}\label{pr1}

		We start by stating a crucial theorem:
		\begin{named}[\textbf{Theorem 4.5.5 of \cite{d19}}]\label{thm3}
			Suppose $B_t$ is adapted to $\Hp_{t}$, and let $p_t=\Prob(B_t|\Hp_{t-1})$. Then as $T\rightarrow\infty$
			\begin{equation*}
			\frac{\frac{1}{T}\sum_{t=1}^T I_{B_t}}{\frac{1}{T}\sum_{t=1}^Tp_t}\rightarrow 1\quad\text{a.s. on}\quad \{\sum_{t=1}^\infty p_t=\infty\}.
			\end{equation*} 
		\end{named}
		If we let $B_t=\{A_t\in B\}$, and note that $p_t=\Prob(A_t\in B|\Hp_{t-1})$, then Theorem 4.5.5 of \cite{d19} implies
		\begin{equation}
		\frac{\frac{1}{T}\sum_{t=1}^T I_{B_t}}{\frac{1}{T}\sum_{t=1}^Tp_t}\rightarrow 1\quad\text{a.s. on}\quad \{\Prob(A^*(\theta^*)\in B|\Hp_{\infty})>0\},\label{e11}
		\end{equation}
		since $\lim_{t\rightarrow\infty} p_t=\Prob(A^*(\theta^*)\in B|\Hp_{\infty})$ a.s. by \eqref{e10} which leads to 
		\begin{equation*}
		\sum_{t=1}^\infty p_t=\infty\quad\text{a.s. on}\quad\{\Prob(A^*(\theta^*)\in B|\Hp_{\infty})>0\}.
		\end{equation*}
		We also point out that \eqref{e10} implies
		\begin{equation}
		\lim_{T\rightarrow\infty}\frac{1}{T}\sum_{t=1}^{T}p_t=\Prob(A^*(\theta^*)\in B|\Hp_{\infty})\quad\text{a.s.}\label{e12}
		\end{equation}
		by Ces\`aro mean, which states that for any real convergent sequence $\{a_t\}_{t=1}^\infty$:
		\begin{equation}
		\lim_{T\rightarrow\infty}\frac{1}{T}\sum_{t=1}^{T}a_t=\lim_{t\rightarrow\infty}a_t.\label{ef1}
		\end{equation}
		By combining \eqref{e11}, and \eqref{e12}, we arrive at:
		\begin{equation}
		\lim_{T\rightarrow\infty}\frac{1}{T}\sum_{t=1}^{T}I_B(A_t)=\Prob(A^*(\theta^*)\in B|\Hp_{\infty})\quad\text{a.s. on}\quad\{\Prob(A^*(\theta^*)\in B|\Hp_{\infty})>0\}.\label{e13}
		\end{equation}
		Equation \eqref{e13} also implies the following almost surely on $\{\Prob(A^*(\theta^*)\in B^c|\Hp_{\infty})>0\}$
		\begin{align}
		&\Prob(A^*(\theta^*)\in B|\Hp_{\infty})=1-\Prob(A^*(\theta^*)\in B^c|\Hp_{\infty})\nonumber\\
		&=1-\lim_{T\rightarrow\infty}\frac{1}{T}\sum_{t=1}^{T}I_B^c(A_t)\label{e14}\\
		&=\lim_{T\rightarrow\infty}\frac{1}{T}\sum_{t=1}^{T}I_B(A_t)\label{e15}
		\end{align}
		where \eqref{e14} follows from \eqref{e13}. By noting $1-I_{B^C}(A_t)=I_B(A_t)$, we conclude \eqref{e15}. Since almost surely either $\Prob(A^*(\theta^*)\in B|\mathcal{F}_{\infty})$, or $\Prob(A^*(\theta^*)\in B^c|\mathcal{F}_{\infty})$ is positive, we finish the proof by combining \eqref{e13}, and \eqref{e15}.
	
	\subsection{Proof of Theorem \ref{thm4}}\label{pr2}

	We first note that for any $t\in\mathbb{Z}^+$, $\Ex[Y_{A^*,t}-Y_{A_t,t}]$ satisfies the following set of equalities:
	\begin{align}
	&\Ex[Y_{A^*,t}-Y_{A_t,t}]\nonumber\\
	&=\Ex[\Ex[Y_{A^*,t}|\theta^*,A^*(\theta^*)]-\Ex[Y_{A_t,t}|\theta^*,A_t]]\label{e19}\\
	&=\Ex[f(\theta^*,A^*(\theta^*))-f(\theta^*,A_t)],\label{e20}
	\end{align}
	where \eqref{e19} follows from the law of total expectation. Finally \eqref{e2} leads to \eqref{e20}. By \eqref{e4} we know
	\begin{equation}
	f(\theta^*,A^*(\theta^*))-f(\theta^*,A_t)\geq 0\label{e21}
	\end{equation}  
	which leads to $\Ex[Y_{A^*,t}-Y_{A_t,t}]\geq 0$ with help of \eqref{e20}.

	Now let $\Theta^i$ be
	\begin{equation*}
	\Theta^i=\cup_{j=1}^i \Theta_j
	\end{equation*}
	where $\Theta_j$ is given in \eqref{e7}. We first prove a usefull lemma:
	\begin{lemma}\label{lem1}
		If for any $i,j\in\mathbb{Z}^+$ $a_i,a_j\in\mathcal{A}$ and $j\geq i$, then
		\begin{equation*}
		I_{(\Theta^j)^c}(\theta^*)\Prob(A^*(\theta^*)=a_i|\Hp_\infty)=0\quad\text{a.s.}
		\end{equation*}
		where $(\Theta^j)^c=\Theta\backslash\Theta^j$.
		\begin{proof}
			We have that
			\begin{align}
			f(\theta^*,A^*(\theta^*))-f(\theta^*,A_t)&\geq(f(\theta^*,A^*(\theta^*))-f(\theta^*,A_t))I_{(\Theta^j)^c}(\theta^*)I_{\{a_i\}}(A_t)\label{e22}\\
			&=(f(\theta^*,A^*(\theta^*))-f(\theta^*,a_i))I_{(\Theta^j)^c}(\theta^*)I_{\{a_i\}}(A_t)\label{e23}
			\end{align}
			where \eqref{e22} follows from \eqref{e21}. We achieve \eqref{e23} by pointing out the domain of $I_{\{a_i\}}(A_t)$. The first thing to note is that 
			\begin{align}
			\Ex[f(\theta^*,A^*(\theta^*))I_{(\Theta^1)^c}(\theta^*)I_{\{a_i\}}(A_t)]
			&=\Ex[\Ex[f(\theta^*,A^*(\theta^*))I_{(\Theta^j)^c}(\theta^*)I_{\{a_i\}}(A_t)|\Hp_{t-1}]]\label{e24}\\
			&=\Ex[\Ex[f(\theta^*,A^*(\theta^*))I_{(\Theta^j)^c}(\theta^*)|\Hp_{t-1}]\Ex[I_{\{a_i\}}(A_t)|\Hp_{t-1}]]\label{e25}\\
			&=\Ex[f(\theta^*,A^*(\theta^*))I_{(\Theta^j)^c}(\theta^*)\Prob(A^*(\theta^*)=a_i|\Hp_{t-1})]\label{e26}
			\end{align}
			where \eqref{e24} follows from the law of total expectation. Conditioned on the past observations $\Hp_{t-1}$, $\theta^*$ and $A_t$ are independent, which proves \eqref{e25}. Further conditioning the term inside the expectation in \eqref{e26} with respect to $\Hp_{t-1}$ and using \eqref{e8} shows \eqref{e26}. Applying a similar analysis, we can also prove that
			\begin{equation}
			\Ex[f(\theta^*,a_i)I_{(\Theta^j)^c}(\theta^*)I_{\{a_i\}}(A_t)]=\Ex[f(\theta^*,a_i)I_{(\Theta^j)^c}(\theta^*)\Prob(A^*(\theta^*)=a_i|\Hp_{t-1})]\label{e27}.
			\end{equation} 
			Ultimately we conclude that
			\begin{align}
			\Ex[Y_{A^*,t}-Y_{A_t,t}]&=\Ex[f(\theta^*,A^*(\theta^*))-f(\theta^*,A_t)]\label{e28}\\
			&\geq \Ex[(f(\theta^*,A^*(\theta^*))-f(\theta^*,a_i))I_{(\Theta^j)^c}(\theta^*)I_{\{a_i\}}(A_t)]\label{e29}\\
			&=\Ex[(f(\theta^*,A^*(\theta^*))-f(\theta^*,a_i))I_{(\Theta^j)^c}(\theta^*)\Prob(A^*(\theta^*)=a_i|\Hp_{t-1})]\label{e30}\\
			&\geq 0.\label{e31}
			\end{align}
			Equation \eqref{e28} is the restatement  of \eqref{e20}. Equation \eqref{e23} leads to \eqref{e29}. Using \eqref{e26} and \eqref{e27} proves \eqref{e30}. Since the term
			\begin{equation*}
			(f(\theta^*,A^*(\theta^*))-f(\theta^*,a_i))\\I_{(\Theta^j)^c}(\theta^*)\Prob(A^*(\theta^*)=a_i|\Hp_{t-1})\geq 0\quad\text{a.s.}
			\end{equation*}
			by \eqref{e4}, we arrive at \eqref{e31}. Consequently \eqref{e30} and \eqref{e31} combined with the assumption in \eqref{e16} leads to
			\begin{equation}
			\lim_{T\rightarrow\infty}\frac{1}{T}\sum_{t=1}^T (\Ex[(f(\theta^*,A^*(\theta^*))-f(\theta^*,a_i))\\I_{(\Theta^j)^c}(\theta^*)\Prob(A^*(\theta^*)=a_i|\Hp_{t-1})])=0\label{e32}.
			\end{equation}
			
			We note that by \eqref{e10}, we have
			\begin{align*}
			&\lim_{t\rightarrow\infty}(f(\theta^*,A^*(\theta^*))-f(\theta^*,a_i))I_{(\Theta^j)^c}(\theta^*)\Prob(A^*(\theta^*)=a_i|\Hp_{t-1})\\&=(f(\theta^*,A^*(\theta^*))-f(\theta^*,a_i))I_{(\Theta^j)^c}(\theta^*)\Prob(A^*(\theta^*)=a_i|\Hp_{\infty})\quad\text{a.s.}
			\end{align*}
			and since 
			\begin{equation*}
			|(f(\theta^*,A^*(\theta^*))-f(\theta^*,a_i))I_{(\Theta^j)^c}(\theta^*)\Prob(A^*(\theta^*)=a_i|\Hp_{t-1})|\leq 2\sup_{a\in\mathcal{A}}|f(\theta^*,a)|
			\end{equation*}
			almost surely for any $t$ with $\sup_{a\in\mathcal{A}}|f(\theta^*,a)|$ being integrable by \eqref{e3}, dominated convergence theorem dictates that
			\begin{align}
			&\lim_{t\rightarrow\infty}\Ex[(f(\theta^*,A^*(\theta^*))-f(\theta^*,a_i))I_{(\Theta^j)^c}(\theta^*)\Prob(A^*(\theta^*)=a_i|\Hp_{t-1})]\nonumber\\
			&=\Ex[(f(\theta^*,A^*(\theta^*))-f(\theta^*,a_i))I_{(\Theta^j)^c}(\theta^*)\Prob(A^*(\theta^*)=a_i|\Hp_{\infty})]\label{e33}
			\end{align}
			Coupling \eqref{e33} with Ces\`aro mean, \eqref{ef1} means that
			\begin{align*}
				&\lim_{T\rightarrow\infty}\frac{1}{T}\sum_{t=1}^T (\Ex[(f(\theta^*,A^*(\theta^*))-f(\theta^*,a_i))I_{(\Theta^j)^c}(\theta^*)\Prob(A^*(\theta^*)=a_i|\Hp_{t-1})]\\
				&=\Ex[(f(\theta^*,A^*(\theta^*))-f(\theta^*,a_i))I_{(\Theta^j)^c}(\theta^*)\Prob(A^*(\theta^*)=a_i|\Hp_{\infty})]
			\end{align*}
			and consequently with the help of \eqref{e32}, we arrive at
			\begin{equation}
			\Ex[(f(\theta^*,A^*(\theta^*))-f(\theta^*,a_i))I_{(\Theta^j)^c}(\theta^*)\Prob(A^*(\theta^*)=a_i|\Hp_{\infty})]=0.\label{e34}
			\end{equation}
			We know the term inside the expectation in \eqref{e34} is almost surely non-negative by \eqref{e4} which in turn shows that
			\begin{equation}
			(f(\theta^*,A^*(\theta^*))-f(\theta^*,a_i))I_{(\Theta^j)^c}(\theta^*)\Prob(A^*(\theta^*)=a_i|\Hp_{\infty})=0\quad\text{a.s.}\label{e35}
			\end{equation}
			However the way we constructed $A^*$ in \eqref{e5} and \eqref{e6} coupled with the fact that $j\geq i$ means that
			\begin{equation*}
			\{\theta\in\Theta|f(\theta,a_i)\geq f(\theta,a)\quad\forall a\in\mathcal{A}\}\cap(\Theta^j)^c=\emptyset,
			\end{equation*}
			and consequently
			\begin{equation}
			f(\theta^*,A^*(\theta^*))-f(\theta^*,a_i)>0\quad\text{on }\{\theta^*\in(\Theta^j)^c\}.\label{e36}
			\end{equation}
			Finally combining \eqref{e35}, and \eqref{e36} leads to
			\begin{equation}
			I_{(\Theta^j)^c}(A^*(\theta^*))\Prob(A^*(\theta^*)=a_i|\Hp_\infty)=0\quad\text{a.s.}\label{e37}
			\end{equation}
			which is the desired result.
		\end{proof} 
	\end{lemma}
	
	We now return to the proof of Theorem \ref{thm4}. To deduce \eqref{e17} we apply induction to show
	\begin{equation*}
	\Prob(A^*(\theta^*)=a_i|\Hp_\infty)=I_{\{a_i\}}(A^*(\theta^*))\quad\text{a.s.}
	\end{equation*}
	for any $i\in\mathbb{Z}^+$ such that $a_i\in\mathcal{A}$. Let $i=j=1$, then Lemma \ref{lem1} implies that
	\begin{equation}
	I_{(\Theta^1)^c}(\theta^*)\Prob(A^*(\theta^*)=a_1|\Hp_\infty)=0\quad\text{a.s.}\label{e38}
	\end{equation}
	and by \eqref{e5} with \eqref{e6} we know $A^*(\theta^*)=a_1$ if and only if $\theta^*\in\Theta^1$ which in turn helps us reformulate \eqref{e38} as
	\begin{equation}
	I_{\{a_1\}^c}(A^*(\theta^*))\Prob(A^*(\theta^*)=a_1|\Hp_\infty)=0\quad\text{a.s.}\label{e39}
	\end{equation}
	However $0\leq\Prob(A^*(\theta^*)=a_1|\Hp_\infty)\leq 1$ almost surely, and $\Prob(A^*(\theta^*)=a_1|\Hp_\infty)$ has to integrate to $\Prob(A^*(\theta^*)=a_1)$. Then \eqref{e39} implies $\Prob(A^*(\theta^*)=a_1|\Hp_\infty)=1$ almost surely on $\{A^*(\theta^*)=a_1\}$, and as a result
	\begin{equation*}
	\Prob(A^*(\theta^*)=a_1|\Hp_\infty)=I_{\{a_1\}}(A^*(\theta^*))\quad\text{a.s.}
	\end{equation*}
	We just proved the first step of the induction process. Now suppose for any $j<i$ where $i\geq 2$, we know that
	\begin{equation}
	\Prob(A^*(\theta^*)=a_j|\Hp_\infty)=I_{\{a_j\}}(A^*(\theta^*))\quad\text{a.s.}\label{e40}
	\end{equation}
	By Lemma \ref{lem1}, we have for $j\leq i$: 
	\begin{equation*}
	I_{(\Theta^i)^c}(\theta^*)\Prob(A^*(\theta^*)=a_j|\Hp_\infty)=0\quad\text{a.s.}
	\end{equation*}
	which in turn by simple addition leads to 
	\begin{equation}
	I_{(\Theta^i)^c}(\theta^*)\Prob(A^*(\theta^*)\in\cup_{j=1}^i\{a_j\}|\Hp_\infty)=0\quad\text{a.s.}\label{e41}
	\end{equation}
	By \eqref{e5}, and \eqref{e6}, we again note that $A^*(\theta^*)\in\cup_{j=1}^i\{a_j\}$ if and only if $\theta^*\in\Theta^i$. Then we have that
	\begin{equation}
	I_{(\cup_{j=1}^i\{a_i\})^c}(A^*(\theta^*))\Prob(A^*(\theta^*)\in\cup_{j=1}^i\{a_j\}|\Hp_\infty)=0\label{e42}
	\end{equation}
	almost surely, and the argument used after \eqref{e39} naturally carries over here to show that
	\begin{equation}
	\Prob(A^*(\theta^*)\in\cup_{j=1}^i\{a_j\}|\Hp_\infty)=I_{\cup_{j=1}^i\{a_i\}}(A^*(\theta^*))\label{e43}
	\end{equation}
	almost surely. Consequently
	\begin{align}
	I_{\{a_i\}}(A^*(\theta^*))&=I_{\cup_{j=1}^i\{a_i\}}(A^*(\theta^*))-I_{\cup_{j=1}^{i-1}\{a_i\}}(A^*(\theta^*))\nonumber\\
	&=\Prob(A^*(\theta^*)\in\cup_{j=1}^i\{a_j\}|\Hp_\infty)-\Prob(A^*(\theta^*)\in\cup_{j=1}^{i-1}\{a_j\}|\Hp_\infty)\label{e44}\\
	&=\Prob(A^*(\theta^*)=a_i|\Hp_\infty)\quad\text{a.s.}\label{e45}
	\end{align}
	where \eqref{e44} directly follows from the induction step \eqref{e40} and \eqref{e43}. This result finishes the proof by induction since \eqref{e45} is the desired result.
	
	Now that we proved 
	\begin{equation}
	\Prob(A^*(\theta^*)=a_i|\Hp_\infty)=I_{\{a_i\}}(A^*(\theta^*))\quad\text{a.s.}\label{e46}
	\end{equation}
	for any $i$, we see that \eqref{e46} leads to \eqref{e17} since $B$ is countable due to $\mathcal{A}$ being countable. 
	
	We finish the proof by noting that \eqref{e10} combined with \eqref{e17} leads to \eqref{e18}.

\bibliography{sample}

\begin{thebibliography}{8}
\providecommand{\natexlab}[1]{#1}
\providecommand{\url}[1]{#1}
\csname url@samestyle\endcsname
\providecommand{\newblock}{\relax}
\providecommand{\bibinfo}[2]{#2}
\providecommand{\BIBentrySTDinterwordspacing}{\spaceskip=0pt\relax}
\providecommand{\BIBentryALTinterwordstretchfactor}{4}
\providecommand{\BIBentryALTinterwordspacing}{\spaceskip=\fontdimen2\font plus
\BIBentryALTinterwordstretchfactor\fontdimen3\font minus
  \fontdimen4\font\relax}
\providecommand{\BIBforeignlanguage}[2]{{%
\expandafter\ifx\csname l@#1\endcsname\relax
\typeout{** WARNING: IEEEtranN.bst: No hyphenation pattern has been}%
\typeout{** loaded for the language `#1'. Using the pattern for}%
\typeout{** the default language instead.}%
\else
\language=\csname l@#1\endcsname
\fi
#2}}
\providecommand{\BIBdecl}{\relax}
\BIBdecl

\bibitem[Thompson(1933)]{ts}
W.~R. Thompson, ``On the likelihood that one unknown probability exceeds
  another in view of the evidence of two samples,'' \emph{Biometrika}, vol.
  25(3/4), pp. 285--294, 1933.

\bibitem[Chapelle and Li(2011)]{cl11}
O.~Chapelle and L.~Li, ``An empirical evaluation of thompson sampling,''
  \emph{Advances in neural information processing systems}, pp. 2249--2257,
  2011.

\bibitem[Russo et~al.(2018)Russo, {Van Roy}, Kazerouni, Osband, and Wen]{dw18}
D.~Russo, B.~{Van Roy}, A.~Kazerouni, I.~Osband, and Z.~Wen, ``A tutorial on
  thompson sampling,'' \emph{Foundations and Trends\textregistered\text{ }in
  Machine Learning}, vol. 11(1), pp. 1--96, 2018.

\bibitem[Russo and {Van Roy}(2014)]{rv14}
D.~Russo and B.~{Van Roy}, ``Learning to optimize via posterior sampling,''
  \emph{Mathematics of Operations Research}, vol. 39(4), pp. 1221--1243, 2014.

\bibitem[Russo and Van~Roy(2016)]{rv16}
D.~Russo and B.~Van~Roy, ``An information-theoretic analysis of thompson
  sampling,'' \emph{The Journal of Machine Learning Research}, vol.~17, no.~1,
  pp. 2442--2471, 2016.

\bibitem[Kaufmann et~al.(2012)Kaufmann, Korda, and Munos]{km12}
E.~Kaufmann, N.~Korda, and R.~Munos, ``Thompson sampling: An asymptotically
  optimal finite-time analysis,'' \emph{Proceedings of the 24th International
  Conference on Algorithmic Learning Theory}, pp. 199--213, 2012.

\bibitem[Agrawal and Goyal(2017)]{ag17}
S.~Agrawal and N.~Goyal, ``Near-optimal regret bounds for thompson sampling,''
  \emph{Journal of the ACM (JACM)}, vol. 64(5), pp. 1--24, 2017.

\bibitem[Durrett(2019)]{d19}
R.~Durrett, \emph{Probability: Theory and Examples}.\hskip 1em plus 0.5em minus
  0.4em\relax Cambridge university press, 2019.

\end{thebibliography}
\bibliographystyle{icml2020}



\end{document}